\documentclass{article}
\usepackage{ijcai24}

\usepackage{times}
\usepackage{soul}
\usepackage{url}
\usepackage[hidelinks]{hyperref}
\usepackage[utf8]{inputenc}
\usepackage[small]{caption}
\usepackage{graphicx}
\usepackage{amsmath}
\usepackage{amsthm}
\usepackage{booktabs}
\usepackage{algorithm}
\usepackage{algorithmic}
\usepackage[switch]{lineno}
\usepackage{longtable}
\usepackage{multirow}
\usepackage{subfigure,hyperref}
\usepackage{makecell}
\usepackage[marginal]{footmisc}

\usepackage{subfigure,float}
\usepackage{amssymb,amsfonts}
\usepackage{color}
\hypersetup{
}

\title{Predictive Accuracy-Based Active Learning for Medical Image Segmentation}

\author{
Jun Shi $^{1}$
\and
Shulan Ruan $^{1}$
\and
Ziqi Zhu$^{2}$
\and
Minfan Zhao$^1$
\and
Hong An$^{1,3}$
\and \\
Xudong Xue$^{4}$
\and
Bing Yan$^{5}$\\
\affiliations
$^1$School of Computer Science and Technology, University of Science and Technology of China\\
$^2$School of Data Science, University of Science and Technology of China \\
$^3$Laoshan Laboratory\\
$^4$Hubei Cancer Hospital, Tongji Medical College, Huazhong University of Science and Technology\\
$^5$Department of Radiation Oncology, The First Affiliated Hospital of USTC, \\Division of Life Sciences and Medicine, University of Science and Technology of China\\
\emails
\{shijun18,~slruan\}@mail.ustc.edu.cn,
han@ustc.edu.cn
}

\begin{document}
\maketitle

\begin{abstract}
    Active learning is considered a viable solution to alleviate the contradiction between the high dependency of deep learning-based segmentation methods on annotated data and the expensive pixel-level annotation cost of medical images. However, most existing methods suffer from unreliable uncertainty assessment and the struggle to balance diversity and informativeness, leading to poor performance in segmentation tasks. In response, we propose an efficient \textbf{P}redictive \textbf{A}ccuracy-based \textbf{A}ctive \textbf{L}earning (\textbf{PAAL}) method for medical image segmentation, first introducing predictive accuracy to define uncertainty. Specifically, PAAL mainly consists of an Accuracy Predictor (\textbf{AP}) and a Weighted Polling Strategy (\textbf{WPS}). The former is an attached learnable module that can accurately predict the segmentation accuracy of unlabeled samples relative to the target model with the predicted posterior probability. The latter provides an efficient hybrid querying scheme by combining predicted accuracy and feature representation, aiming to ensure the uncertainty and diversity of the acquired samples. 
    Extensive experiment results on multiple datasets demonstrate the superiority of PAAL.
    PAAL achieves comparable accuracy to fully annotated data while reducing annotation costs by approximately \textbf{50\%} to \textbf{80\%}, showcasing significant potential in clinical applications. The code is available at \url{https://github.com/shijun18/PAAL-MedSeg}.
\end{abstract}

\section{Introduction}

Recently, supervised deep learning methods have been widely applied to medical image segmentation tasks, such as delineating organs and lesions \cite{wang2022medical}. Despite the remarkable potential exhibited by current methods, the inherent data-hungry nature leads to their superior performance being heavily reliant on large-scale annotated data, posing a major challenge in real-world clinical scenarios, as the pixel-wise annotation of medical images is experience-dependent and labor-intensive \cite{jiao2023learning}. To address this problem, researchers have devoted considerable efforts to exploring various data-efficient methods \cite{feng2021interactive,ren2021survey,jiao2023learning,zhang2023dive} to achieve higher segmentation performance. As an iterative learning method, Active Learning (AL) can actively select {\em the most valuable or informative} samples for annotation during the training process, the purpose of which is to use as little annotated data as possible to achieve optimal model performance \cite{zhan2022comparative}. As a result, AL is particularly applicable to medical image segmentation, characterized by high annotation costs and difficulty.

\begin{figure}[]
	\centering
	\includegraphics[width=1\columnwidth]{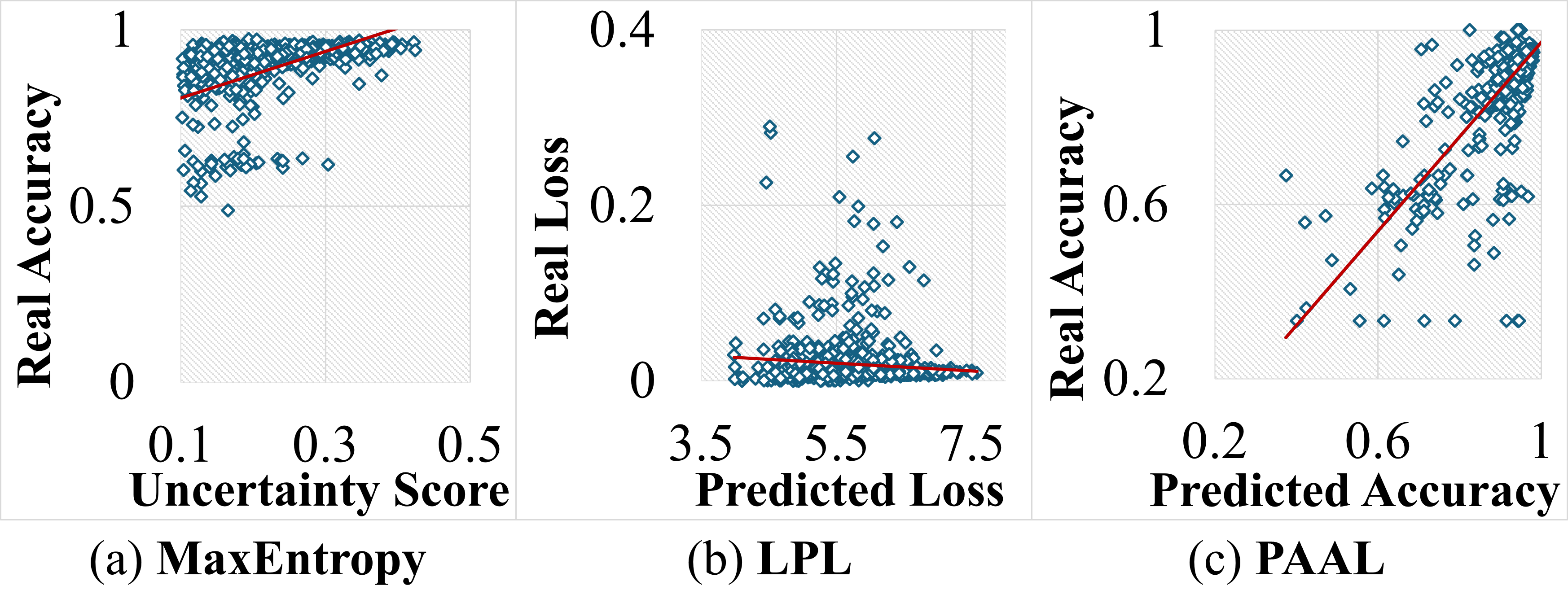} 
	\caption{Data visualization of different methods. We use the model from the last active learning cycle to obtain the uncertainty scores, predicted loss, and predicted accuracy. 750 images chosen from the ACDC dataset are shown. The red line is the fitted line. }
	\label{uncer-acc}
\end{figure}

Existing deep AL methods (pool-based) can be categorized into three branches \cite{zhan2022comparative}: uncertainty-based, diversity-based, and combined strategies. The core idea of uncertainty-based methods is to query and annotate those samples with high uncertainty. The typical methods \cite{li2013adaptive,joshi2009multi,brinker2003incorporating,wang2014new,kampffmeyer2016semantic} utilize the predicted posterior probability of the target model to measure uncertainty. However, the overconfidence of deep neural networks often leads to unreliable uncertainty assessments \cite{zhan2022comparative}. 
As shown in Figure \ref{uncer-acc}(a), the uncertainty scores arising from the Maximum Entropy approach \cite{li2013adaptive} fail to reflect the segmentation accuracy of the current model on unlabeled samples. 
Some studies \cite{li2020attention,yang2017suggestive,kim2023active} adopt the bootstrapping strategy to enhance the uncertainty assessment, by leveraging the disagreement of the online committee. AB-UNet \cite{saidu2021active} exploits the Dropout mechanism to emulate a Bayesian network and compute uncertainty according to the Monte Carlo average of multiple forward passes. Despite the performance gain, these methods suffer from significantly increased computational costs, unsuitable for deeper networks and larger datasets. Besides, LPL \cite{yoo2019learning} first proposes a task-agnostic loss prediction strategy to directly predict the loss as the uncertainty of unlabeled samples relative to the target model. However, LPL introduces a joint optimization problem and completely ignores the importance of the predicted posterior probability for uncertainty assessment, resulting in limited performance. 
Figure \ref{uncer-acc}(b) illustrates that the segmentation loss predicted by the LPL method is highly inconsistent with the actual loss.

Diversity-based approaches aim to query samples that provide varied information for annotation, and most of them use two clustering methods: KMeans \cite{bodo2011active} and CoreSet \cite{sener2018active}. KMeans-based methods perform unsupervised clustering on unlabeled samples according to the intermediate features of the target model and then select those samples closest to each centroid while CoreSet-based approaches construct a representative subset as a proxy for the entire dataset. These methods can boost the sample diversity but tend to overlook the informativeness of the acquired samples. Consequently, they are often deemed complementary to uncertainty-based methods, giving rise to a series of combined querying strategies. For instance, Exploration-Exploitation \cite{yin2017deep} builds upon the Maximum Entropy strategy and integrates a determinantal point process to select the most uncertain and diverse samples. BADGE \cite{ash2019deep} proposes a two-stage querying approach. The first stage forms a coarse candidate set based on gradient embedding, while the second stage refines the candidate set through KMeans++ clustering. Other combined strategies \cite{zhou2017fine,shui2020deep} are also designed to uphold both the informativeness and diversity of the selected samples, yet their high complexity entails another engineering cost. Therefore, the primary challenge faced by AL methods in medical image segmentation is: { \em how to more accurately and cost-effectively assess uncertainty while maintaining a balance in the diversity of the selected samples.}

To address these challenges, our initial focus is optimizing uncertainty assessment to overcome the shortcomings of existing methods in terms of accuracy and computational efficiency. Inspired by LPL \cite{yoo2019learning}, we design a predictive accuracy-driven uncertainty assessment method. The motivation behind it is: {\em if it is possible to predict the loss of a sample point, then why not predict its accuracy relative to the target model?} Our preliminary experiments have demonstrated the feasibility of the accuracy prediction, as shown in Figure  \ref{uncer-acc}(c). 
The predicted accuracy of our proposed method exhibits good consistency with the actual accuracy. To this end, we propose a Predictive Accuracy-based Active Learning  (\textbf{PAAL})  method for medical image segmentation, first introducing the concept of accuracy prediction. The core idea of PAAL is to use a trained lightweight network to predict the segmentation accuracy of the target model on unlabeled samples, and then guide a diversity-based querying strategy to ensure both uncertainty and diversity of the selected samples. 

Specifically, our PAAL mainly consists of an Accuracy Predictor (\textbf{AP}) and a Weighted Polling Strategy (\textbf{WPS}). The AP is a simple neural network that takes the image and the corresponding model predictions as input, aiming to minimize the difference between the predicted and real accuracy. Notably, we design an end-to-end framework to support the simultaneous training of the segmentation model and the attached AP while decoupling their optimization processes. Compared to LPL, our proposed method avoids the joint optimization problem and can leverage the posterior probability to guide the accuracy prediction. Based on this, we design WPS to balance the uncertainty and diversity of the queried samples. After the unsupervised clustering, WPS converts the predicted accuracy of each sample into query weight and cyclically queries the sample with the highest weight in each cluster until iteration ends.
Moreover, we propose an Incremental Querying (IQ) mechanism to ensure training stability and facilitate achieving higher performance under a fixed budget.
In summary, our contributions mainly include:


\begin{itemize}
\item  We first propose the concept of Accuracy Predictor (\textbf{AP}) and design a novel active learning method (\textbf{PAAL}) for medical image segmentation. By using the posterior probability as a guide, the attached AP achieves a high consistency between the predicted and actual accuracy, enabling a more accurate measurement of uncertainty.

\item  We propose a hybrid Weighted Polling Strategy (\textbf{WPS}) to balance the uncertainty and diversity of the acquired samples. Compared to existing methods, our method realizes higher accuracy and more diversified sample distribution, effectively mitigating the issue of imbalanced inter-class annotation.

\item  
Extensive experimental results prove that PAAL outperforms existing methods, achieving accuracy comparable to fully annotated data while reducing annotation costs by approximately \textbf{50\%} to \textbf{80\%}.

\end{itemize}

\section{Related Work}

\subsection{Active Learning}

Active Learning (AL) aims to minimize annotation costs and maximize model performance by selecting the most informative samples for annotation.
In this paper, we discuss pool-based active learning methods, which access multiple samples at once. Given an unlabeled sample pool, three main approaches are utilized: uncertainty-based, diversity-based, and combined strategies \cite{zhan2022comparative}. Among them, uncertainty-based methods \cite{li2013adaptive,wang2014new,yuval2011reading,kampffmeyer2016semantic} typically use the posterior probability predicted by the target model to define uncertainty. For instance, the Maximum Entropy approach \cite{li2013adaptive} selects those samples with the highest prediction entropy. Due to the overconfidence of deep neural networks, the uncertainty estimation of such methods is often unreliable. Some studies optimize uncertainty assessment by adopting the bootstrapping strategy \cite{beluch2018power} or simulating the Bayesian system \cite{gal2017deep,kendall2017uncertainties} while introducing higher engineering costs. Diversity-based methods use the intermediate features of the network for unsupervised clustering of samples. These methods can identify the most representative sample points but ignore the informativeness of the selected samples and are thus often considered complementary to uncertainty-based methods. The combined strategies \cite{yin2017deep,ash2019deep} desire to balance the diversity and uncertainty of the acquired samples and have become the major research direction of AL.

\subsection{AL for Medical Image Segmentation}

Early AL methods are primarily used for image classification. Recently, many researchers have explored the applications of AL methods in medical image segmentation. Due to the differences in network output, most uncertainty-based methods require specific adjustments for segmentation tasks. In contrast, diversity-based methods apply to any task and network since they depend on intermediate features rather than the task-specific output. BioSegment \cite{rombaut2022biosegment} develops a framework that extends typical AL methods to medical image segmentation tasks. In particular, Li et al. \cite{li2020attention} combine the uncertainty assessment based on bootstrapping strategy with similarity representation, proposing a multi-stage combined query strategy. AB-UNet \cite{saidu2021active} adds multiple Dropout layers into the segmentation network to simulate a Bayesian network. It calculates sample uncertainty by obtaining Monte Carlo averages of multiple forward passes. Besides, some studies \cite{cai2021revisiting,saidu2019medical} introduce the concept of super-pixel, which decomposes annotation query from image-level to region-level, attempting to control annotation costs more finely. Other works \cite{blanch2017cost,zhao2021dsal} unite the advantages of AL and semi-supervised learning, using high-confidence pseudo-labels to enhance model performance. 

\section{Methodology }

A more accurate uncertainty assessment leads to better performance of AL \cite{zhan2022comparative}. As a result,  most existing methods for medical image segmentation explore solutions to enhance uncertainty assessment, such as using a bootstrapping strategy \cite{li2020attention} or Bayesian network \cite{saidu2021active}. However, these methods exhibit limited performance while significantly increasing computational complexity. The underlying reason is that uncertainty estimation based on the posterior probability can be negatively affected by the overconfidence of the network, as segmentation predictions often contain considerable noise, especially during the early stages of training. LPL \cite{yoo2019learning} suggests utilizing neural networks to model the mapping between the hidden features of images and actual loss. While LPL has shown performance gains in classification tasks, it fails to accurately predict the dense prediction loss based solely on image features and introduces a convergence issue of multi-objective optimization. Inspired by this, we propose a Predictive Accuracy-based Active Learning (\textbf{PAAL}) method as an alternative learnable uncertainty assessment solution, desiring to overcome the limitations of LPL via simple yet effective designs, as shown in Figure \ref{overview}. 

\subsection{Problem Definition}

Given an arbitrary medical image segmentation task, let $\mathcal{D}$, $\mathcal{D}_{u}$, $\mathcal{D}_{l}$, and $\mathcal{B}$ to represent the entire dataset, the unlabeled data pool, the labeled data set, and the specified annotation budget (simply referring to the maximum number of labeled samples), respectively. Following the standard setup of pool-based active learning methods, we have an initial labeled set $\mathcal{D}_{l} = \{(\mathbf{x}_{i},\mathbf{y}_{i})\}_{i=1}^{M}$ and a large-scale pool of unlabeled samples $\mathcal{D}_{u} = \{\mathbf{x}_{i}\}_{i=1}^{N}$, where $M \ll N$, and $\mathbf{x}_i$ and $\mathbf{y}_i$ represent the $i$-th image and its corresponding true segmentation mask. In the $t$-th iteration of the proposed method, firstly, based on the current segmentation model $\mathcal{M}^{t}$, accuracy predictor $\mathcal{P}^{t}$, and query strategy $\alpha_{wps}=(\mathcal{D}_{u},\mathcal{M}^{t},\mathcal{P}^{t})$, a subset $\mathcal{D}_{q}^{t}$ of batch size $b$ is selected from $\mathcal{D}_{u}$, where $b = \lfloor{\mathcal{B} / T} \rfloor$ and $T$ is the pre-set maximum number of iterations that varies with the annotation budget. Then,  we directly query their true labels from the oracle to construct the labeled subset $\mathcal{D}_{q}^{t,*} = \{(\mathbf{x}_{i},\mathbf{y}_{i})\}_{i=1}^{b}$,  simulating human annotation. Finally, we update $\mathcal{D}_{u}=\mathcal{D}_{u} \setminus \mathcal{D}_{q}^{t}$ and $\mathcal{D}_{l} = \mathcal{D}_{l} \cup \mathcal{D}_{q}^{t,*}$, and retrain $\mathcal{M}$ and $\mathcal{P}$ using $\mathcal{D}_{l}$. The iteration process terminates when the budget $\mathcal{B}$ is exhausted, and the network converges to a stable state. In this paper, our goal is to maximize the segmentation accuracy of the model using as little labeled data as possible. Since the quality of $\mathcal{D}_{l}$ is positively related to the performance of $\mathcal{M}$, the key to this study lies in optimizing the query function $\alpha_{wps}$.

\subsection{Overview Architecture}

\begin{figure*}[]
	\centering
	\includegraphics[width=2\columnwidth]{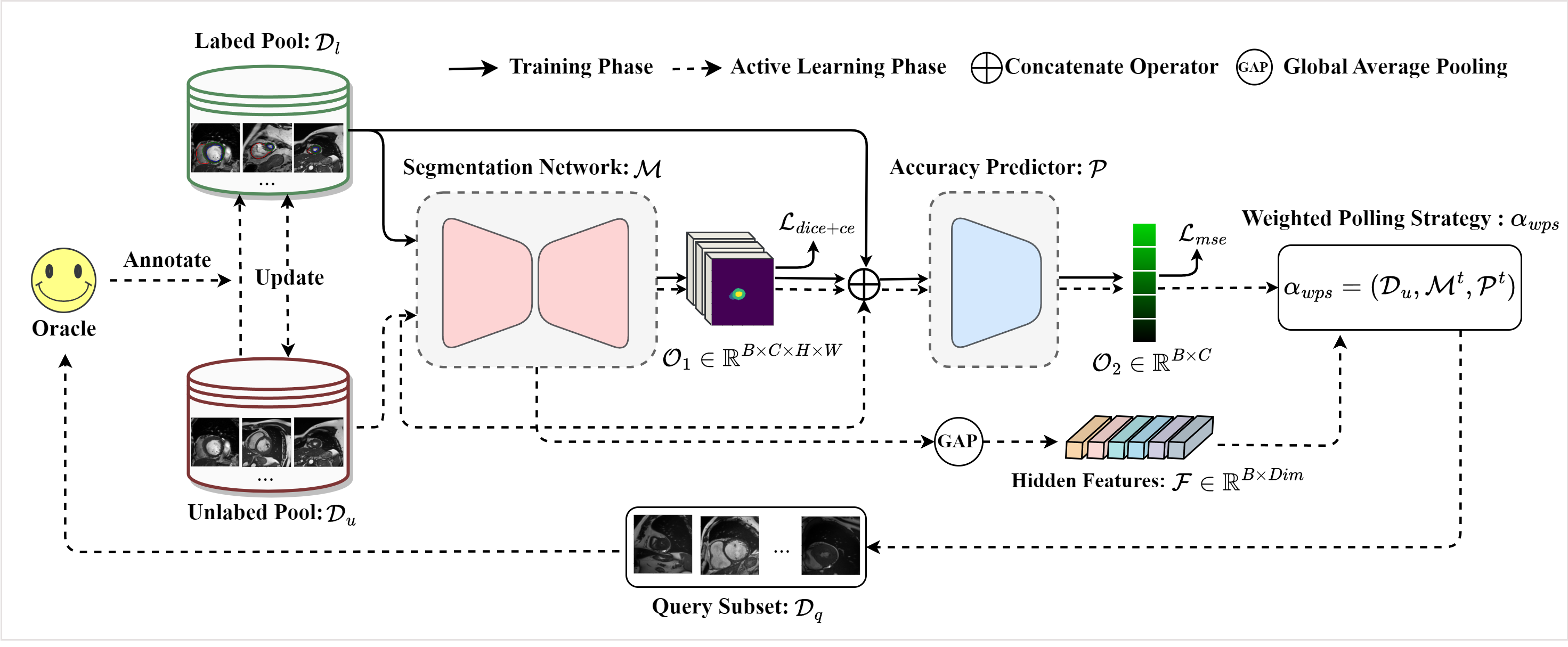} 
	\caption{Overview of our proposed PAAL, where $Dim = 2048$, $t$ denotes the $t$-th iteration, $\mathcal{L}_{dice + ce}$ represents the combined loss.}
	\label{overview}
\end{figure*}

Figure \ref{overview} illustrates the overall structure and workflow of PAAL. In addition to the basic segmentation network $\mathcal{M}$, PAAL includes two main modules: an Accuracy Predictor $\mathcal{P}$ (\textbf{AP}) and a Weighted Polling Strategy $\alpha_{wps}$  (\textbf{WPS}). The former predicts the segmentation accuracy of the target model for unlabeled samples, while the latter selects a subset $\mathcal{D}_{q}$ with the most informative and diverse. Without loss of generality, we utilize the U-Net \cite{ronneberger2015u} with an encoder of ResNet-50 \cite{he2016deep} as the base model in the experiments. Unlike LPL, we completely decouple the optimization processes of the AP and the segmentation network, embedding them into an end-to-end unified training framework in a cascaded manner. Besides, our AP utilizes the posterior probability of the segmentation network as prior information, aiming to minimize the discrepancy between the predicted and actual accuracy. More importantly, we design the WPS that utilizes both the predicted accuracy and feature representations of samples to balance the uncertainty and diversity of the acquired samples. The following sections provide detailed explanations of these two core modules.

\subsection{Accuracy Predictor}

The Accuracy Predictor (\textbf{AP}) of PAAL is fundamentally a regression model in the form of a deep neural network. The primary consideration involves the selection of a suitable accuracy metric as the regression target. Multiple metrics are available for quantifying segmentation accuracy, such as the Dice Similarity Coefficient (DSC), Hausdorff Distance, and pixel-wise classification accuracy. Considering the stability of convergence, we empirically select a metric with a value range within $[0,1]$ to represent ``{\em Predictive Accuracy}". We use DSC as the regression target in the experiments, although it can be replaced with any normalized metric. To reduce computation, AP is a simple variant of ResNet-18, which adds a Sigmoid layer behind the linear layer of the classification head. During training, we concatenate the input image $\mathcal{O}_{0} \in \mathbb{R}^{B\times C' \times H \times W}$ and its corresponding segmentation prediction probability $\mathcal{O}_{1} \in \mathbb{R}^{B\times C \times H \times W}$  along the channel dimension and deliver it to AP, where $B$,  $C'$, $H$, $W$, and $C$ denote the batch size, number of image channels, height and width of the image, and the number of segmentation categories, respectively. The optimization process utilizes Mean Squared Error (MSE) loss $\mathcal{L}_{mse}$ to minimize the difference between the predicted $\mathcal{O}_{2} \in \mathbb{R}^{B\times C}$ and actual accuracy. Notably, during the early stages of joint training, we set a brief silent period (5 epochs) for AP, meaning that only the segmentation network undergoes training during this period to alleviate the impact of early segmentation noise. Compared to the loss prediction module of LPL, the optimization process of the proposed AP is independent of the segmentation network, thereby bypassing the multi-objective optimization problem. Further, utilizing the posterior probability of the segmentation network, rather than solely relying on image features, as prior information helps to enhance convergence. Experimental results showcase a high consistency between predicted and actual accuracy, affirming the effectiveness of the proposed accuracy prediction approach.

\subsection{Weighted Polling Strategy}
From the perspective of data efficiency, the pixel-level annotation of medical images faces two challenges: sample redundancy and imbalanced inter-class annotation. Taking 3D images as an example, a CT or MR dataset typically contains numerous highly similar slices due to the similarity of anatomical structures, leading to redundant annotations and reduced data efficiency. Moreover, there are significant volume differences between tissues or organs, and small-volume targets only appear in a few slices, causing an imbalanced annotation distribution that may hamper the segmentation accuracy of the model for minority classes. Although AP can identify the samples with high uncertainty, it fails to ensure their diversity. To address these issues, we propose a hybrid Weighted Polling Strategy (\textbf{WPS}) to balance the informativeness and diversity of the selected samples.

\begin{algorithm}[h]
    \caption{The Proposed PAAL Process }
    \label{alg:algorithm}
    \textbf{Input}: Unlabeled dataset $\mathcal{D}_{u}$, Initial labeled dataset $\mathcal{D}_{l}$ , Segmentation network $\mathcal{M}$, Accuracy predictor $\mathcal{P}$, Oracle  \\
    \textbf{Parameter}: Maximum iterations $T$, Number of clusters $K$\\
    \textbf{Output}: Final $\mathcal{D}_{l}$, $\mathcal{M}$ and $\mathcal{P}$ 
    \begin{algorithmic}[1] 
        \STATE $t \leftarrow 1$,    $IQ \leftarrow 0$   
        \WHILE{not reach the budget and stable convergence}
        \STATE $\mathcal{M}, \mathcal{P} \leftarrow \mathcal{D}_{l}$
        \IF {$t \le T$ and IQ $\geq$ 10}
        \STATE $\mathcal{O}_{1}, \mathcal{F} \leftarrow (\mathcal{M}, \mathcal{D}_{u})$ 
        \STATE $\mathcal{O}_2 \leftarrow (\mathcal{P}, \mathcal{O}_{1}, \mathcal{D}_{u})$ 
        \STATE $\mathcal{W} \leftarrow \mathcal{O}_{2}$    
        \STATE $\{\Omega_i\}_{i=1}^{K} \leftarrow $ ($\mathcal{F},K$)      
        \STATE $\mathcal{D}_{q}^{t} \leftarrow$ ($\{\Omega_i\}_{i=1}^{K}, \mathcal{W}$)  
        \STATE $\mathcal{D}_{q}^{t,*} \leftarrow$  Oracle($\mathcal{D}_{q}^{t}$) 
        \STATE $\mathcal{D}_{l} \leftarrow \mathcal{D}_{l} \cup \mathcal{D}_{q}^{t,*}$  
        \STATE $\mathcal{D}_{u} \leftarrow \mathcal{D}_{u} \setminus \mathcal{D}_{q}^{t}$  
        \STATE $t \leftarrow t + 1$, $IQ \leftarrow 0$
        \ENDIF
        \IF { not $\mathcal{M} \uparrow$}
        \STATE $IQ \leftarrow IQ + 1$
        \ELSE
        \STATE $IQ \leftarrow 0$
        \ENDIF
        \ENDWHILE
        \STATE \textbf{return} $\mathcal{D}_{l}$, $\mathcal{M}$ and $\mathcal{P}$
    \end{algorithmic} 
\end{algorithm}

As illustrated in Algorithm \ref{alg:algorithm}, in the query process of active learning, WPS first transforms the predicted accuracy of unlabeled samples into query weights $\mathcal{W}$. The query weight $w_i$ for the $i$-th sample is negatively correlated with the overall predicted accuracy. The computation is detailed as follows:
\begin{flalign}
\mathcal{W}=\{{w}_{i}\}_{i=1}^{N},  {w}_{i} = \frac{1}{C} \sum_{j=1}^{C} -log(p^{j}_{i}),
\end{flalign}
where $p_{i}^{j}$ denotes the predicted accuracy for the $j$-th segmentation class of the $i$-th sample. We employ the logarithmic mean to amplify attention to minority classes. Then, based on the hidden features of the segmentation model, a naive KMeans algorithm is utilized for unsupervised clustering of the unlabeled samples, yielding $K$ clusters $\{\Omega_i\}_{i=1}^{K}$. To reduce computational complexity, we set a small $K = \lfloor log_{2}(b*4) + 1 \rfloor$ and use an adaptive global average pooling layer to compress the original representations. Finally, we alternately query the sample with the highest weight in each cluster until the current iteration concludes. Compared to existing diversity-based methods, WPS exhibits lower computational complexity, suitable for deeper networks and larger datasets, and considers both the uncertainty and distribution of the selected samples to alleviate the issue of imbalanced inter-class annotation. Besides, we propose an Incremental Querying (\textbf{IQ}) mechanism that differs from querying based on specified epochs. We design a simple trigger mechanism, initiating the next query only when the current model fails to achieve a performance gain over ten consecutive epochs. This ensures training stability and facilitates achieving higher performance within a fixed budget.

\section{Experiments and Results}

\subsection{Datasets}

As shown in Table \ref{tab:datasets}, datasets used in our experiments include (1) Brain Tumour \cite{antonelli2022medical}: a multi-modal Magnetic Resonance (MR) dataset provided by the Medical Segmentation Decathlon (MSD), comprising 484 annotated samples with segmentation targets of brain Edema, Enhanced (ET) and non-Enhanced tumors (nET); 
(2) SegTHOR \cite{lambert2020segthor}: a chest Computed Tomography (CT) dataset containing only 40 scans, annotated for 4 organs;
(3) ACDC \cite{bernard2018deep}: a commonly used cardiac MR dataset composed of scan images from 100 patients, annotated for the Left Ventricle (LV), Right Ventricle (RV), and Myocardium (Myo).
More importantly, to explore the application potential of the proposed method, we constructed (4) Liver OAR: a clinical organ-at-risk (OAR) segmentation dataset for liver cancer, annotated for 8 abdominal organs, collected by the Radiotherapy Department of the First Affiliated Hospital of University of Science and Technology of China, where all CT images were annotated and verified by two experienced physicists and have been used in radiotherapy planning. We set different initial annotation ratios for different datasets due to the varying slice scales.

Specifically, we list the annotation ratio for different categories on different datasets as follows. 
For Brain Tumour, (Edma, nET, ET) = (50\%, 35\%, 31\%);
for SegTHOR, (Esophagus, Heart, Trachea, Aorta) = (53\%, 21\%, 27\%, 51\%);
for ACDC, (RV, Myo, LV) = (82\%, 95\%, 95\%);
and for Liver OAR, (Spinal-Cord, Small-Intestine, Kidney-L, Kidney-R, Liver, Heart, Lung-L, Lung-R) = (80\%, 44\%, 22\%, 20\%, 30\%, 17\%, 34\%, 34\%). We can see that except for the ACDC dataset, the other datasets suffer from different degrees of inter-class annotation imbalance. During training, each dataset was split into training and validation sets at a ratio of 8:2 for five-fold cross-validation. All reported DSC results in subsequent sections are the average and standard deviation of the five-fold.

\begin{table}[t]
    \centering
    \scalebox{0.95}{\begin{tabular}{lccc}
        \toprule
        Dataset  & Modality & Samples (slices)   & Init.R.\\
        \midrule
        Brain Tumour    & multi-MR      & 484 (66,512)     & 0.5\%  \\
        SegTHOR         & CT            & 40 (7,420)       & 5.0\% \\
        ACDC            & MR            & 100 (1,902)      & 5.0\% \\
        Liver OAR *   & CT            & 49 (3,725)       & 5.0\% \\
        \bottomrule
    \end{tabular}}
    \caption{Datasets used in the experiments. * denotes the private dataset from clinical, and Init.R. refers to the initial annotation ratio.}
    \label{tab:datasets}
\end{table}

\begin{table*}[]
\scalebox{0.8}{
\begin{tabular}{lccc|ccc|ccc}
\hline
\multirow{2}{*}{Method} & \multicolumn{3}{c}{ACDC}           & \multicolumn{3}{c}{SegTHOR}           & \multicolumn{3}{c}{Brain Tumour}           \\  [2pt]\cline{2-10}
                        & 10\%          & 20\%            & 50\%          & 10\%          & 20\%            & 50\%           & 5\%          & 10\%            & 20\%         \\  [2pt] \hline
\emph{Random}                  & {{ 85.3$\pm$2.9}} & {{ 87.9$\pm$2.9}} & {{ 90.3$\pm$1.5}} & {{ 81.1$\pm$1.2}} & {{ 84.6$\pm$0.8}} & {{ 86.4$\pm$0.7}} & {{ 70.7$\pm$4.2}} & {{ 71.7$\pm$2.6}} & {{ 73.5$\pm$2.9}} \\ [2pt] \hline
MaxEntropy \cite{li2013adaptive}                & 83.0$\pm$2.3   & 86.6$\pm$3.3 & 89.8$\pm$2.0   & 75.1$\pm$7.1 & 81.9$\pm$1.0   & 85.9$\pm$1.4  & 67.3$\pm$4.8 & 68.7$\pm$4.6 & 71.1$\pm$3.8  \\ [2pt]
LeastConf \cite{wang2014new}        & 83.0$\pm$3.5   & 86.2$\pm$3.7 & 89.7$\pm$1.9 & 78.8$\pm$1.4 & 82.0$\pm$1.1 & 85.3$\pm$1.1 & 68.6$\pm$5.1 & 69.3$\pm$4.0 & 72.2$\pm$2.9 \\ [2pt]
VarRatio \cite{zhan2022comparative}               & 82.8$\pm$2.5 & 85.1$\pm$3.4 & 89.4$\pm$2.4 & 75.4$\pm$8.2 & 78.6$\pm$8.3 & 84.9$\pm$1.2 & 68.2$\pm$4.9 & 69.9$\pm$3.8 & 71.3$\pm$3.4 \\
Margin \cite{yuval2011reading}                  & 84.6$\pm$2.9 & 85.9$\pm$3.9 & 89.4$\pm$2.0   & 76.6$\pm$5.7 & 78.7$\pm$8.6 & 85.5$\pm$1.6   & 65.8$\pm$7.2 & 69.7$\pm$4.1 & 71.4$\pm$2.8   \\
KMeans \cite{rombaut2022biosegment}                  & 84.6$\pm$4.3 & 86.0$\pm$2.5   & 90.2$\pm$1.6 & 81.2$\pm$1.9 & 84.8$\pm$0.8 & 86.6$\pm$0.9 & 71.3$\pm$4.2 & 73.1$\pm$2.0 & 73.3$\pm$4.2 \\
CoreSet \cite{zhan2022comparative}                 & 85.0$\pm$2.6   & 87.4$\pm$1.9 & 90.3$\pm$1.5 & — & — & — & — & — & —  \\
Entropy+KMeans \cite{yin2017deep}         & 82.8$\pm$4.0   & 86.6$\pm$3.6 & 89.8$\pm$2.2 & 79.1$\pm$2.7 & 84.8$\pm$0.5 & 86.4$\pm$0.9 & 69.8$\pm$4.3 & 70.7$\pm$4.4 & 72.8$\pm$2.9 \\
AB-UNet \cite{saidu2021active}                & 82.2$\pm$3.4 & 86.9$\pm$2.1 & 90.2$\pm$1.4 & 81.3$\pm$1.3 & 84.7$\pm$0.8 & 86.4$\pm$0.6 & 71.5$\pm$3.4 & 72.6$\pm$2.8 & 73.4$\pm$2.5 \\
CEAL \cite{blanch2017cost}                    & 83.5$\pm$2.4 & 86.2$\pm$2.9 & 89.5$\pm$2.3 & 70.6$\pm$8.5 & 77.9$\pm$7.6 & 84.7$\pm$1.3 & 67.3$\pm$5.8 & 70.2$\pm$3.0 & 71.3$\pm$3.5  \\
LPL \cite{yoo2019learning}                     & 70.9$\pm$5.9  & 80.2$\pm$3.5 & 87.6$\pm$3.2 & 75.4$\pm$2.1 & 78.2$\pm$1.3 & 83.6$\pm$1.1 & 51.5$\pm$8.6 & 61.7$\pm$4.0 & 66.6$\pm$4.7 \\ \hline
PAAL (only AP)               & 86.3$\pm$2.5 & 89.1$\pm$2.0 & 90.7$\pm$1.2 & 82.8$\pm$1.2 & 85.5$\pm$0.2 & 86.9$\pm$0.9 & 71.7$\pm$0.8 & 72.9$\pm$1.2 & 73.9$\pm$1.1 \\
\textbf{PAAL}        & \textbf{86.8$\pm$2.2} & \textbf{89.5$\pm$1.3} & \textbf{91.1$\pm$1.5} & \textbf{84.3$\pm$1.3} & \textbf{85.7$\pm$0.5} & \textbf{87.5$\pm$0.6} & \textbf{72.2$\pm$1.7} & \textbf{74.0$\pm$2.0} & \textbf{75.6$\pm$1.1} \\ \hline
\emph{Full data}               & \multicolumn{3}{c|}{{\color{red} 91.6$\pm$1.4}}           & \multicolumn{3}{c|}{{\color{red} 88.5$\pm$1.3}}           & \multicolumn{3}{c}{{\color{red} 76.4$\pm$1.7}}     \\    \hline
\end{tabular}
}

\caption{
Comparison with state-of-the-art methods on 3 open-source datasets under different annotation ratios. We show the mean$\pm$std (standard deviation) of DSC (\%) score for five-fold cross-validation. Bold is the best result, and — denotes that the method is not applicable.
}
\label{tab:overall}
  
\end{table*}

\subsection{Implementation Details}

PAAL and all baselines are implemented using PyTorch and integrated into a unified training framework.
In particular, we select representative methods as baselines, including uncertainty-based, diversity-based, and combined methods, all of which have open-source implementations. All models are trained from scratch on 8 NVIDIA A800 GPUs, with the same loss function, e.g. the combined loss \cite{shi2023rethinking} of Dice and Cross-Entropy for segmentation model and MSE loss for AP.
We set 3 maximum querying ratios for different datasets according to varying slice scales: $\{5\%, 10\%, 20\%\}$ for the Brain Tumour dataset and $\{10\%, 20\%, 50\%\}$ for the other datasets. Unlike the proposed IQ mechanism, the query interval for comparison methods is set to 5 epochs. The maximum iterations for each dataset are related to the maximum querying ratio. Specifically, Brain Tumour dataset has maximum iterations of $\{10, 15, 15\}$, while the other datasets have the same $\{5, 15, 20\}$. For the Brain Tumour dataset, the slice resolution is resized to $4 \times 256 \times 256$, while for the other datasets, it is $1 \times 512 \times 512$. We employ AdamW optimizer \cite {loshchilov2018decoupled} with an initial learning rate of 1e-3, a batch size of 64, and use the cosine annealing strategy \cite {loshchilov2016sgdr} to control the learning rate, with a weight decay of 1e-4, warm-up epochs of 10, and the minimum learning rate of 1e-6. Each model is evaluated on the validation set at the end of every epoch. To alleviate overfitting, we adopt an early stopping strategy with a tolerance of 40 epochs to search for the best model within 400 epochs and apply data augmentation, including random distortion, rotation, flip, and noise.

\subsection{Overall Performance}

\paragraph{Results on Open-Source Datasets.} In Table \ref{tab:overall}, we report the results of the proposed PAAL on open-source single-modal datasets ACDC \cite{bernard2018deep} and SegTHOR \cite{lambert2020segthor}, as well as the multi-modal dataset Brain Tumour \cite{antonelli2022medical}, compared with state-of-the-art methods. PAAL (only AP) indicates the removal of the WPS module, selecting samples solely based on query weights, similar to other uncertainty-based methods. We can see that PAAL significantly outperforms all previous methods, achieving the highest DSC across different datasets with varying annotation ratios. In particular, at the lowest annotation budget, our proposed method surpasses typical Maximum Entropy \cite{li2013adaptive}, KMeans \cite{rombaut2022biosegment}, and LPL \cite{yoo2019learning} methods by \textbf{3.8\%}, \textbf{2.2\%}, and \textbf{15.9\%} on ACDC, \textbf{9.2\%}, \textbf{3.1\%}, and \textbf{8.9\%} on SegTHOR, and \textbf{4.9\%}, \textbf{0.9\%}, and \textbf{20.7\%} on Brain Tumour, respectively. These results demonstrate the superior data efficiency of PAAL under a limited budget. Notably, the performance differences among different methods diminish as the annotation ratio increases. PAAL achieves segmentation accuracy comparable to fully annotated data at annotation ratios of \textbf{50\%} and \textbf{20\%}, respectively.

An intriguing observation is that most uncertainty-based methods relying on the posterior probability perform even worse than random sampling, implying their limited applicability to segmentation tasks. We hypothesize that this phenomenon stems from the potential noise introduced by network overconfidence, making uncertainty assessment prone to failure in dense prediction tasks. The experimental results provide supporting evidence. For example, AB-UNet \cite{saidu2021active} uses a Bayesian network to enhance uncertainty assessment and achieves the performance gain, while CEAL \cite{blanch2017cost} using pseudo-labels performs worse in most cases, indicating the low reliability of the network predictions. Although diversity-based methods \cite{rombaut2022biosegment,zhan2022comparative} can maintain relatively satisfactory performance, they are constrained by network depth and data scale. Moreover, LPL performs poorly due to convergence issues arising from joint optimization, especially on the complicated Brain Tumour datasets. In contrast, our proposed PAAL outperforms all comparison methods even using AP alone, demonstrating its effectiveness.

\begin{table}[t]
\scalebox{0.8}{\begin{tabular}{lcccc}
\hline
\multirow{2}{*}{Method} & \multicolumn{3}{c}{Liver OAR}   & \multirow{2}{*}{Query Time}        \\  [2pt]\cline{2-4}
                                     & 10\%          & 20\%            & 50\%        \\  [2pt] \hline
\emph{Random}                        &  {{ 86.5$\pm$5.3}} &  {{ 89.8$\pm$0.5}} & {{ 91.4$\pm$0.4}}  & — \\ \hline
MaxEntropy        &  80.0$\pm$9.0 & 88.9$\pm$0.9 & 91.4$\pm$0.4  & 13.57 \\
LeastConf       &  86.4$\pm$1.2 & 88.9$\pm$0.6 & 91.4$\pm$0.6 & 13.26 \\
VarRatio  &  83.0$\pm$8.4 & 89.0$\pm$0.6 & 91.1$\pm$0.5  &  13.86 \\
Margin     &  87.4$\pm$0.6 & 89.3$\pm$0.6 & 91.4$\pm$0.6  & 13.41\\
KMeans  &  88.0$\pm$0.7 & 89.7$\pm$0.4 & 91.3$\pm$0.6  & 21.40 \\
CoreSet   &  86.3$\pm$5.0 & 90.1$\pm$0.5 & 91.4$\pm$0.4  & 47.88 \\
Entropy+KMeans   &  87.6$\pm$1.5 & 89.5$\pm$0.7 & 91.0$\pm$0.4  & 22.75 \\
AB-UNet   &  88.1$\pm$1.1 & 90.1$\pm$0.1 & 91.1$\pm$0.5 & 240.67  \\
CEAL          &  85.0$\pm$4.0 & 88.2$\pm$0.4 & 90.8$\pm$0.4 & 26.65 \\
LPL           &  86.5$\pm$1.0 & 88.4$\pm$0.9 & 90.4$\pm$0.8  & 13.55  \\  \hline
PAAL (only AP)                               &  89.2$\pm$0.8 & 90.4$\pm$0.3 & 91.4$\pm$0.5  &  \textbf{12.68} \\
\textbf{PAAL}                  &\textbf{89.7$\pm$0.4} & \textbf{90.8$\pm$0.6} & \textbf{91.9$\pm$0.2}  & 20.24  \\  \hline

\emph{Full data}      & \multicolumn{3}{c}{{\color{red} 92.3$\pm$1.6}}  & —    \\    \hline
\end{tabular}}

\caption{DSC (\%) score and average query time (s) on the private dataset of different methods. — denotes without AL process.}
\label{tab:liver}

\end{table}

\begin{table}[t]
\scalebox{1}{\begin{tabular}{lccc}
\hline
\multirow{2}{*}{Method} & \multicolumn{3}{c}{ACDC}           \\  [2pt]\cline{2-4}
                                     & 10\%          & 20\%            & 50\%        \\  [2pt] \hline
\emph{Random}                        & {{ 85.3$\pm$2.9}} & {{ 87.9$\pm$2.9}} & {{ 90.3$\pm$1.5}}\\ [2pt] \hline
w/o WPS and IQ & 85.3$\pm$2.5 & 88.5$\pm$2.1 & 90.7$\pm$1.1 \\
w/o IQ  & 86.4$\pm$2.6 & 89.4$\pm$1.5 & 90.9$\pm$1.4\\
w/o WPS & 86.3$\pm$2.5 & 89.1$\pm$2.0 & 90.7$\pm$1.2  \\
w/o AP  & 84.6$\pm$4.3 & 86.0$\pm$2.5 & 90.2$\pm$1.6 \\
\textbf{PAAL}                   & \textbf{86.8$\pm$2.2} & \textbf{89.5$\pm$1.3} & \textbf{91.1$\pm$1.5}  \\ \hline
\emph{Full data}      & \multicolumn{3}{c}{{\color{red} 91.6$\pm$1.4}}      \\    \hline
\end{tabular}}
\caption{Ablation study on ACDC dataset, w/o denotes without.}
\label{tab:ab}
\end{table}

\paragraph{Results on Private Clinical Dataset.} Table \ref{tab:liver} shows the results on the small-scale private dataset, Liver OAR. Similarly, PAAL achieves the highest DSC at 10\%, 20\%, and 50\% annotation ratios, reaching \textbf{89.7\%}, \textbf{90.8\%}, and \textbf{91.9\%}, respectively. Notably, at an annotation budget of 20\%, except for our proposed method, CoreSet, and AB-UNet, all other comparative methods perform worse than random sampling. Given the computational complexity, CoreSet is unsuitable for deeper networks and larger datasets, and AB-UNet also requires additional computations to simulate the Bayesian network. In contrast, PAAL achieves higher segmentation performance with lower computational overhead, especially when using only AP. These results further validate the effectiveness of PAAL in reducing annotation costs, showcasing significant potential in practical applications.

\begin{figure}[t]
	\centering
	\includegraphics[width=1\columnwidth]{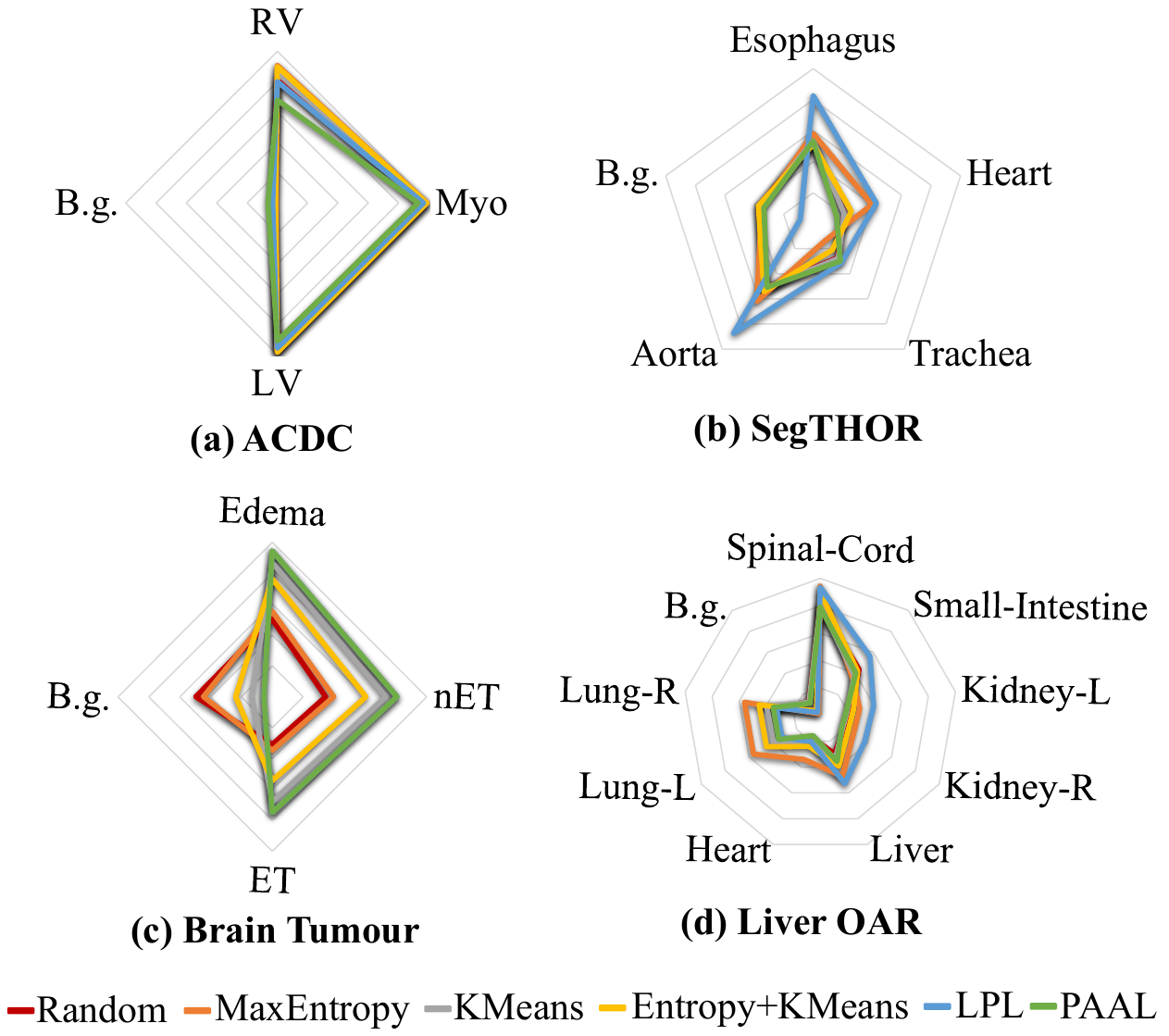} 
	\caption{The annotation distribution of different methods.}
	\label{qa}
\end{figure}

\paragraph{Time Efficiency Analysis.} The average query time in Table \ref{tab:liver} demonstrates the time efficiency of different methods. The query time of PAAL mainly consists of the inference time of AP and KMeans clustering time of the WPS module. It's evident that by reducing the feature dimension and the cluster size of the KMeans method, the query time of PAAL is still higher than that of the uncertainty-based methods, but lower than that of the diversity-based methods. After WPS removal, the time efficiency of our proposed method is better than that of all comparison methods due to the lightweight structure of AP. Overall, PAAL achieves a good trade-off in terms of accuracy and time efficiency.

\begin{figure}[t]
	\centering
	\includegraphics[width=1\columnwidth]{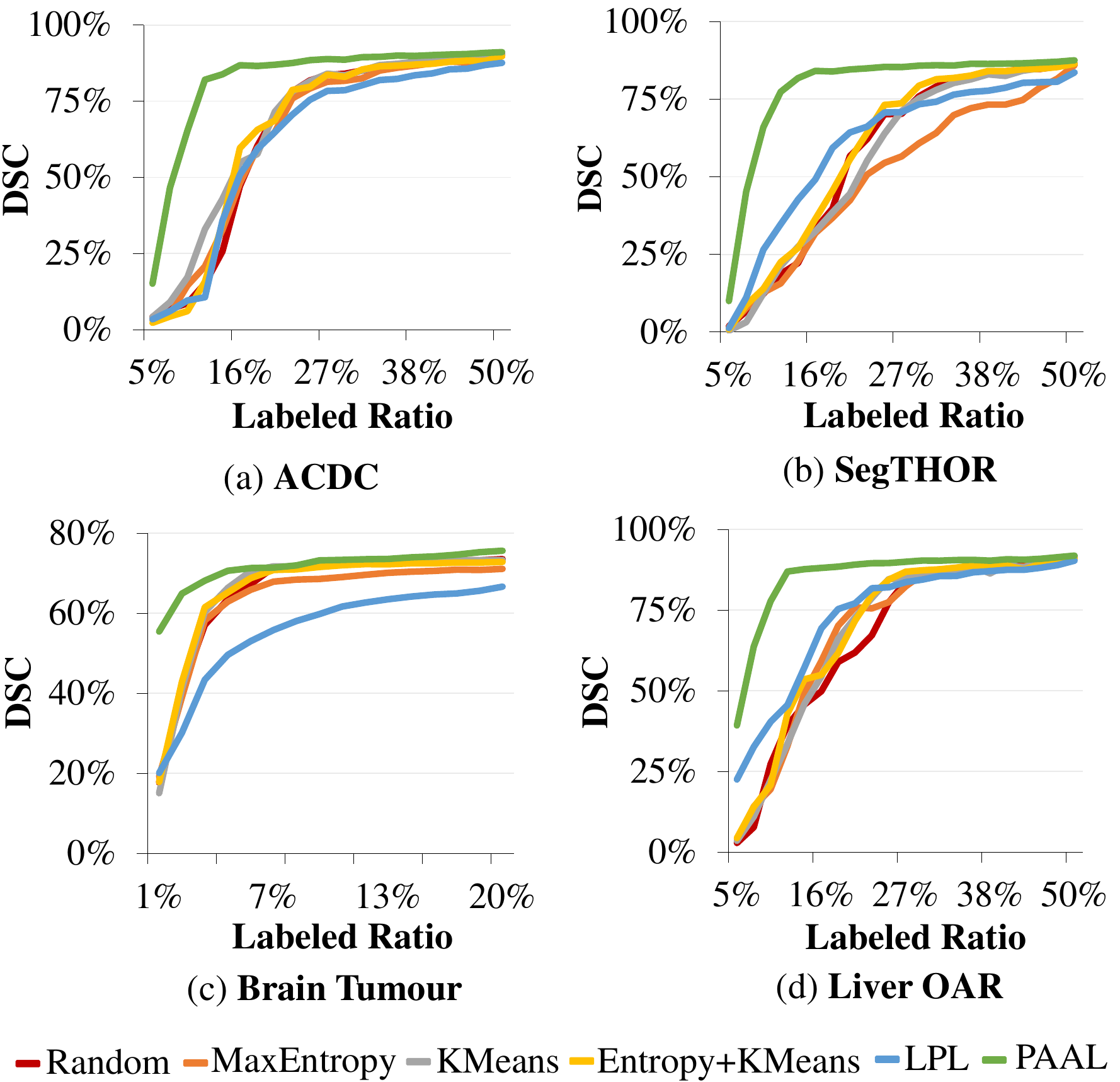} 
	\caption{The performance curves under different labeled ratios.}
	\label{curve}
\end{figure}

\subsection{Ablation Study}

To reveal the effect of different modules on performance improvement, we conduct ablation studies on AP, WPS, and IQ modules on the ACDC dataset, and the results are shown in Table \ref{tab:ab}. After removing each module separately, there is a varying degree of performance decline, indicating the essential importance of all modules for PAAL. It is worth noting that w/o AP means that PAAL degenerates into the KMeans method, where the most significant performance degradation occurs. In low-budget scenarios, the IQ leads to a notable improvement in overall performance, demonstrating the effectiveness of IQ in enhancing network convergence. Furthermore, we can observe that both AP and WPS have a significant impact on overall performance, further highlighting the superiority of our designs.

\subsection{Quantitative Analysis}

As shown in Figure \ref{qa}, we compare the annotation distribution of samples selected by different query strategies under the maximum annotation budget (20\% for Brain Tumour, 50\% for the others). Notably, we introduce a background category denoted as B.g., to represent those images without any segmentation target. It can be observed that, compared to the original distribution represented by Random, different methods exhibit significant differences. For the ACDC and Brain Tumour datasets, PAAL achieves better performance by significantly increasing the annotation ratio of minority classes or reducing the ratio of majority classes. For example, the annotated slice ratio of “Edema” and “ET” is reduced from 1.62 to 1.26. As for the other two datasets, the annotation distribution of PAAL is generally consistent with the original distribution, while LPL and Maximum Entropy show significant differences, leading to poor performance. These results demonstrate that PAAL can adaptively adjust based on the data distribution of the specific task, helping to alleviate the problem of imbalanced annotation of multiple categories.

Furthermore, we present the performance curves of different query strategies as the annotation ratio iteratively increases under the maximum annotation budget. As shown in Figure \ref{curve}, unlike the drastic fluctuations of the existing methods, the DSC rising curve of the proposed PAAL is remarkably smooth.
The essential reason is that the proposed IQ mechanism ensures training stability by triggering queries only after achieving the best performance on current data.
Besides, PAAL consistently maintains the best segmentation performance under different annotation ratios, further proving its effectiveness and superiority.

\section{Conclusion}

In this paper, we proposed a Predictive Accuracy-based Active Learning (\textbf{PAAL}) approach for medical image segmentation.
Specifically, we employed a lightweight Accuracy Predictor (\textbf{AP}) to directly predict the segmentation accuracy of unlabeled samples related to the target model, and designed a hybrid Weighted Polling Strategy (\textbf{WPS}) to balance uncertainty and diversity. Extensive experimental results demonstrated the superiority of PAAL over existing methods.
The low complexity and high data efficiency of PAAL indicated significant potential for clinical applications.
In the future, we will explore more optimization methods such as semi-supervised learning to further enhance the performance.




\section*{Acknowledgments}

This work is financially supported by the National Key Research and Development Program of China (Grants No. 2016YFB1000403), the Fundamental Research Funds for the Central Universities (Grant No. YD2150002001), and the National Natural Science Foundation of China (Grant No. 60970023).

\section*{Contribution Statement}
Jun Shi and Shulan Ruan have equal contributions to this paper. Hong An is the corresponding author.

\bibliographystyle{named}
\bibliography{mybib}

\end{document}